%% file: lrpaper.tex
\documentclass{article} 
\usepackage{iclr2016_conference,times}
\usepackage[colorlinks=true]{hyperref}
\hypersetup{
  urlcolor     = purple, 
  linkcolor    = blue, 
  citecolor   = blue 
}
\usepackage{url}
\usepackage{times,epsfig,graphicx,amsmath,amssymb,subcaption,color,multirow,booktabs}
\usepackage{siunitx}
\usepackage{pgfplotstable}
\usepackage[compact]{titlesec}
\titlespacing{\subsection}{0pt}{1ex}{0ex}
\titlespacing{\subsubsection}{0pt}{0.5ex}{0ex}
\usepackage{pdflscape}

\input{ACsettings}
\input{pgftablehighlight}

\title{Training CNNs with Low-Rank Filters for\\Efficient Image Classification}

\author{
Yani Ioannou$^{1}$, Duncan Robertson$^{2}$, Jamie Shotton$^{2}$, Roberto Cipolla$^1$ \& Antonio Criminisi$^2$\\
$^1$University of Cambridge, $^2$Microsoft Research\\
{\footnotesize\texttt{\{yai20,rc10001\}@cam.ac.uk, \{a-durobe,jamiesho,antcrim\}@microsoft.com }}
}

\iclrfinalcopy 

\begin{document}

\maketitle
\begin{abstract} 
We propose a new method for creating computationally efficient convolutional neural networks (CNNs) by using low-rank representations of convolutional filters.
Rather than approximating filters in previously-trained networks with more efficient versions, we learn a set of small basis filters from scratch; during training, the network learns to combine these basis filters into more complex filters that are discriminative for image classification. To train such networks, a novel weight initialization scheme is used. This allows effective initialization of connection weights in convolutional layers composed of groups of differently-shaped filters. We validate our approach by applying it to several existing CNN architectures and training these networks from scratch using the CIFAR, ILSVRC and MIT Places datasets. Our results show similar or higher accuracy than conventional CNNs with much less compute. Applying our method to an improved version of VGG-11 network using global max-pooling, we achieve comparable validation accuracy using 41\% less compute and only 24\% of the original VGG-11 model parameters; another variant of our method gives a 1 percentage point {\em increase} in accuracy over our improved VGG-11 model, giving a top-5 \emph{center-crop} validation accuracy of 89.7\% while reducing computation by 16\% relative to the original VGG-11 model. Applying our method to the GoogLeNet architecture for ILSVRC, we achieved comparable accuracy with 26\% less compute and 41\% fewer model parameters. Applying our method to a near state-of-the-art network for CIFAR, we achieved comparable accuracy with 46\% less compute and 55\% fewer parameters. 
\end{abstract}


\input{introduction}

\input{method}

\input{training}

\input{results}

\input{conclusion}

{\small
\bibliography{lrpaper}
\bibliographystyle{iclr2016_conference}
}
\clearpage
{\huge\textsc{Appendices}}
\appendix
\input{initialization}
\input{mavstimings}
\input{bigpicture}
\input{extraplots}
\input{vggmodeltable}
\end{document}

%% file: ACsettings.tex
\definecolor{accolornotes}{rgb}{0.7,0.3,0.2}
\newcommand{\acnote}[1]{\textcolor{accolornotes}{}}

\definecolor{changes}{rgb}{0.45,0,0}

\definecolor{yicolornotes}{rgb}{0.3,0.7,0.2}
\newcommand{\yinote}[1]{\textcolor{yicolornotes}{}}

\definecolor{jscolornotes}{rgb}{0.7,0.3,0.7}
\newcommand{\js}[1]{\textcolor{jscolornotes}{}}

\definecolor{acgray}{rgb}{0.8,0.8,0.8}

\definecolor{myred}{rgb}{0.6, 0, 0}
\definecolor{myblue}{rgb}{0.3, 0.1, 0.9}

\newcommand{\eg}{{\it e.g.}\ }
\newcommand{\ie}{{\it i.e.}\ }

\newcommand{\vs}{{v.s.}\ }

\newcommand{\be}{\begin{equation}}
\newcommand{\ee}{\end{equation}}
\newcommand{\bea}{\begin{eqnarray}}
\newcommand{\eea}{\end{eqnarray}}
\newcommand{\beas}{\begin{eqnarray*}}
\newcommand{\eeas}{\end{eqnarray*}}

\renewcommand{\eqref}[2][\reflabel]{(\ref{eq:#1-#2})} 

\newcommand{\reflabel}{dummy} 

\DeclareMathAlphabet{\mathcal}{OMS}{cmsy}{m}{n}

%% file: pgftablehighlight.tex
\newcommand{\findmax}[3]{
    \pgfplotstablevertcat{\datatable}{#1}
    \pgfplotstablecreatecol[
    create col/expr={%
    \pgfplotstablerow
    }]{rownumber}\datatable
    \pgfplotstablesort[sort key={#2},sort cmp={float >}]{\sorted}{\datatable}%
    \pgfplotstablegetelem{0}{rownumber}\of{\sorted}%
    \pgfmathtruncatemacro#3{\pgfplotsretval}
    \pgfplotstableclear{\datatable}
}

\newcommand{\findmin}[3]{
    \pgfplotstablevertcat{\datatable}{#1}
    \pgfplotstablecreatecol[
      create col/expr={%
    \pgfplotstablerow
    }]{rownumber}\datatable
    \pgfplotstablesort[sort key={#2},sort cmp={float <}]{\sorted}{\datatable}%
    \pgfplotstablegetelem{0}{rownumber}\of{\sorted}%
    \pgfmathtruncatemacro#3{\pgfplotsretval}
    \pgfplotstableclear{\datatable}
}

\pgfplotstableset{
    highlight col max/.code 2 args={
        \findmax{#1}{#2}{\maxval}
        \edef\setstyles{\noexpand\pgfplotstableset{
                every row \maxval\noexpand\space column #2/.style={
                    postproc cell content/.append style={
                        /pgfplots/table/@cell content/.add={$\noexpand\bf}{$}
                    },
                }
            }
        }\setstyles
    },
    highlight col min/.code 2 args={
        \findmin{#1}{#2}{\minval}
        \edef\setstyles{\noexpand\pgfplotstableset{
                every row \minval\noexpand\space column #2/.style={
                    postproc cell content/.append style={
                        /pgfplots/table/@cell content/.add={$\noexpand\bf}{$}
                    },
                }
            }
        }\setstyles
    },
    highlight row max/.code 2 args={
        \pgfmathtruncatemacro\rowindex{#2-1}
        \pgfplotstabletranspose{\transposed}{#1}
        \findmax{\transposed}{\rowindex}{\maxval}
        \edef\setstyles{\noexpand\pgfplotstableset{
                every row \rowindex\space column \maxval\noexpand/.style={
                    postproc cell content/.append style={
                        /pgfplots/table/@cell content/.add={$\noexpand\bf}{$}
                    },
                }
            }
        }\setstyles
    },
    highlight row min/.code 2 args={
        \pgfmathtruncatemacro\rowindex{#2-1}
        \pgfplotstabletranspose{\transposed}{#1}
        \findmin{\transposed}{\rowindex}{\maxval}
        \edef\setstyles{\noexpand\pgfplotstableset{
                every row \rowindex\space column \maxval\noexpand/.style={
                    postproc cell content/.append style={
                        /pgfplots/table/@cell content/.add={$\noexpand\bf}{$}
                    },
                }
            }
        }\setstyles
    },
}

\makeatletter
\long\def\pgfplotstabletypeset@opt@collectarg[#1]#2{%

    \pgfplotstable@isloadedtable{#2}%
        {\pgfplotstabletypeset@opt@[#1]{#2}}%
        {\pgfplotstabletypesetfile@opt@[#1]{#2}}%
}
\makeatother

%% file: introduction.tex
\section{Introduction}
Convolutional neural networks (CNNs) have been used increasingly succesfully to solve challenging computer vision problems such as image classification \citep{NIPS2012_4824}, object detection \citep{ren2015noc}, and human pose estimation \citep{Tompson_2015_CVPR}.
However, recent improvements in recognition accuracy have come at the expense of increased model size and computational complexity.
These costs can be prohibitive for deployment on low-power devices, or for fast analysis of videos and volumetric medical images.

\begin{figure}[t] 
\centerline{
\includegraphics[width=0.61\columnwidth]{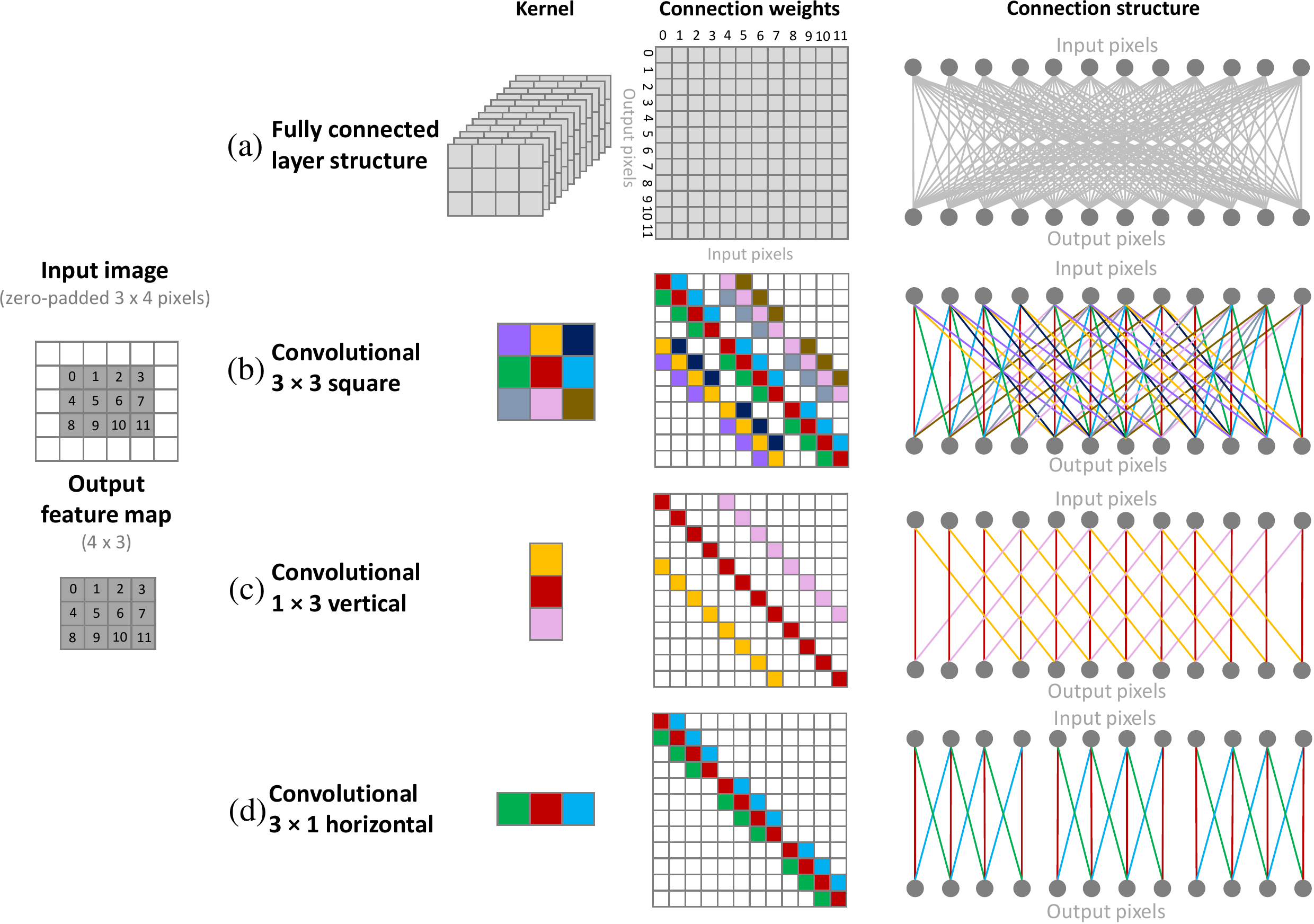}
}
\caption[Image access map visualizing sparsity of convolutional filters.]{
{\bf Network connection structure for convolutional layers.} An input image is transformed in one layer of an neural network into an output image of same size. Connection weight maps show pairwise dependencies between input and output pixels. In (a), each node is connected to all input pixels. For (b,c,d), output pixels depend only on a subset of input pixels (shared weights are represented by unique colours). Note that sparsity increases from (a) to (d), opening up potentially more efficient implementation.
}
\label{fig:sparseconn}
\end{figure}

One promising aspect of CNNs, from a memory and computational efficiency standpoint, is their use of \emph{convolutional filters} (a.k.a.~kernels). Such filters usually have limited spatial extent and their learned weights are shared across the image spatial domain to provide translation invariance~\citep{fukushima:neocognitronbc,lecun-gradientbased-learning-applied-1998}.
Thus, as illustrated in Fig.~\ref{fig:sparseconn}, in comparison with fully connected network layers (Fig.~\ref{fig:sparseconn}a), convolutional layers have a much sparser connection structure and use fewer parameters (Fig.~\ref{fig:sparseconn}b).
This leads to faster training and test, better generalization, and higher accuracy.

This paper focuses on reducing the computational complexity of the convolutional layers of CNNs by further sparsifying their connection structures.  Specifically, we show that by representing convolutional filters using a basis space comprising groups of filters of different spatial dimensions (examples shown in Fig.~\ref{fig:sparseconn}c and d), we can significantly reduce the computational complexity of existing state-of-the-art CNNs without compromising classification accuracy.

Our contributions include a novel method of learning a set of small basis filters that are combined to represent larger filters efficiently. Rather than approximating previously trained networks, we train networks \emph{from scratch} and show that our convolutional layer representation can improve both efficiency and classification accuracy. We further describe how to initialize connection weights effectively for training networks with composite convolutional layers containing groups of differently-shaped filters, which we found to be of critical importance to our training method.

\subsection{Related Work}
\label{relatedwork}
There has been much previous work on increasing the test-time efficiency of CNNs. Some promising approaches work by making use of more hardware-efficient representations. For example \citet{1502.02551v1} and \citet{vanhoucke2011improving} achieve training- and test-time compute savings by further quantization of network weights that were originally represented as 32 bit floating point numbers. However, more relevant to our work are approaches that depend on new network connection structures, efficient approximations of previously trained networks, and learning low rank filters. 

\paragraph{Efficient Network Connection Structures.}
There has been shown to be significant redundancy in the trained weights of CNNs~\citep{denil2013predicting}. \citet{lecun1989optimal} suggest a method of pruning unimportant connections within networks. However this requires repeated network re-training and may be infeasible for modern, state-of-the-art CNNs requiring weeks of training time. \citet{journals/corr/LinCY13} show that the geometric increase in the number and dimensions of filters with deeper networks can be managed using low-dimensional embeddings. The same authors show that global average-pooling may be used to decrease model size in networks with fully connected layers. \citet{journals/corr/SimonyanZ14a} show that stacked filters with small spatial dimensions (\eg $3\times 3$), can operate on the effective receptive field of larger filters (\eg $5 \times 5$) with less computational complexity.

\paragraph{Low-Rank Filter Approximations.}
\citet{conf/cvpr/RigamontiSLF13} approximate {\em previously trained} CNNs with low-rank filters for the semantic segmentation of curvilinear structures within volumetric medical imagery. They discuss two approaches: enforcing an $L_1$-based regularization to learn approximately low rank filters, which are later truncated to enforce a strict rank, and approximating a set of pre-learned filters with a tensor-decomposition into many rank-1 filters. Neither approach learns low rank filters directly, and indeed the second approach proved the more successful.

The work of \citet{journals/corr/JaderbergVZ14} also approximates the existing filters of previously trained networks. They find separable 1D filters through an optimization minimizing the reconstruction error of the already learned full rank filters. They achieve a 4.5$\times$ speed-up with a loss of accuracy of 1\% in a text recognition problem. However since the method is demonstrated only on text recognition, it is not clear how well it would scale to larger data sets or more challenging problems. A key insight of the paper is that filters can be represented by low rank approximations not only in the spatial domain but also in the channel domain.

Both of these methods show that, at least for their respective applications, low rank approximations of full-rank filters learned in convolutional networks can increase test time efficiency significantly. However, being approximations of pre-trained networks, they are unlikely to improve test accuracy, and can only increase the computational requirements during training.

\paragraph{Learning Separable Filters.}
\citet{mamalet2012simplifying} propose training networks with separable filters on the task of digit recognition with the MNIST dataset. They train networks with \emph{sequential} convolutional layers of horizontal and vertical 1D filters, achieving a speed-up factor of 1.6$\times$, but with a relative increase in test error of 13\% (1.45\% \vs 1.28\%).Our approach generalizes this, allowing both horizontal and vertical 1D filters (and other shapes too) at the same layer and avoiding issues with ordering.  We also demonstrate a decrease in error and on more challenging datasets.

%% file: method.tex
\section{Using Low-Rank Filters in CNNs}

\begin{figure}[t] 
\centering
\begin{subfigure}[l]{0.38\columnwidth}
   \includegraphics[width=\columnwidth, page=1]{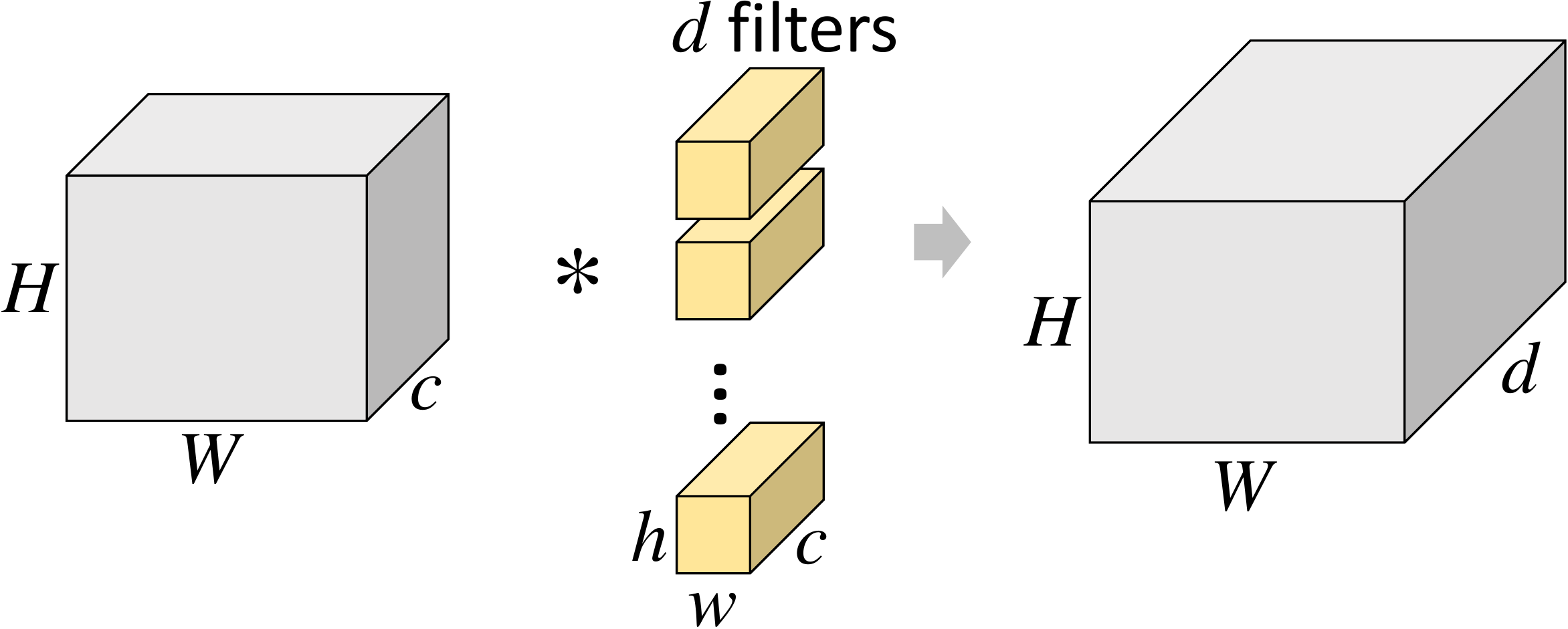}
   \caption{A full rank convolutional layer.}
   \label{fig:fullrank}
\end{subfigure}\\
\begin{subfigure}[b]{0.6\columnwidth}
   \includegraphics[width=\columnwidth, page=2]{figs/sparsification}
   \caption{Sequential separable filters \citep{journals/corr/JaderbergVZ14}.}
   \label{fig:separableseq}
\end{subfigure}\\
\begin{subfigure}[b]{0.6\columnwidth}
   \includegraphics[width=\columnwidth, page=3]{figs/sparsification}
   \caption{Our method, a learned basis space of filters that are rectangular in the spatial domain and oriented horizontally and vertically.}
   \label{fig:ourmethod}
\end{subfigure}\\
\begin{subfigure}[b]{0.6\columnwidth}
   \includegraphics[width=\columnwidth, page=4]{figs/sparsification}
   \caption{Our method, a learned basis space of vertical/horizontal rectangular filters and square filters. Filters of other shapes are also possible.}
   \label{fig:ourmethodfullrank}
\end{subfigure}
\caption[Overview of methods of using low-rank filters.]{{\bf Methods of using low-rank filters in CNNs}. The activation function is not shown, coming after the last layer in each configuration.} 
\label{fig:separablemethods}
\end{figure}

\subsection{Convolutional Filters}

The convolutional layers of a CNN produce output `images' (usually called {\em feature maps}) by convolving input images with one or more learned filters. 
In a typical convolutional layer, as illustrated in Fig.~\ref{fig:fullrank}, a $c$-channel input image of size $H \times W$ pixels is convolved with $d$ filters of size $h \times w \times c$ to create a $d$-channel output image. Each filter is represented by $h w c$ independent weights. Therefore the computational complexity for the convolution of the filter with a $c$-channel input image is $\mathcal{O}(d w h c)$ (per pixel in the output feature map).

In what follows, we describe schemes for modifying the architecture of the convolutional layers so as to reduce computational complexity. The idea is to replace full-rank convolutional layers with modified versions that represent the same number of filters by linear combinations of basis vectors, \ie as lower rank representations of the full rank originals.

\subsection{Sequential Separable Filters}
\label{seqsep}
An existing scheme for reducing the computational complexity of convolutional layers \citep{journals/corr/JaderbergVZ14} is to replace each one with a sequence of two regular convolutional layers but with filters that are rectangular in the spatial domain, as shown in Fig \ref{fig:separableseq}. The first convolutional layer has $m$ filters of size $w \times 1 \times c$, producing an output feature map with $m$ channels. The second convolutional layer has $d$ filters of size $1 \times h \times m$, producing an output feature map with $d$ channels. By this means the full rank original convolutional filter bank is represented by a low rank approximation formed from a linear combination of a set of separable $w \times h$ basis filters.  The computational complexity of this scheme is $\mathcal{O}(m c w)$ for the first layer of horizontal filters and $\mathcal{O}(d m h)$ for the second layer of vertical filters, with a total of $\mathcal{O}(m(c w + d h))$.

Note that \citet{journals/corr/JaderbergVZ14} use this scheme to approximate existing full rank filters belonging to previously trained networks using a retrospective fitting step. In this work, by contrast,  we {\em train} networks containing convolutional layers with this architecture from scratch. In effect, we learn the separable basis filters and their combination weights simultaneously during network training.

\subsection{Filters as Linear Combinations of Bases}
In this work we introduce another scheme for reducing convolutional layer complexity. This works by representing convolutional filters as linear combinations of basis filters as illustrated in Fig.~\ref{fig:ourmethod}. This scheme uses \emph{composite layers} comprising several sets of filters where the filters in each set have different spatial dimensions (see Fig.~\ref{fig:compositelayers}). The outputs of these basis filters may be combined in a subsequent layer containing filters with spatial dimensions $1 \times 1$.

\begin{figure}[tb] 
\centering
\begin{tabular}[c]{rl}
&
\subcaptionbox{$3\times1$ filters.\label{fig:horizontalfilters}}
{
   \includegraphics[width=0.3\columnwidth]{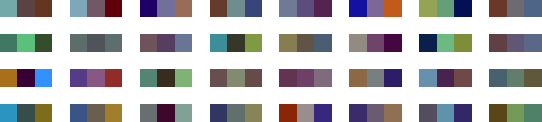}
}\\
\subcaptionbox{$1\times3$ filters.\label{fig:verticalfilters}}[0.3\columnwidth]
{
   \includegraphics[height=0.3\columnwidth]{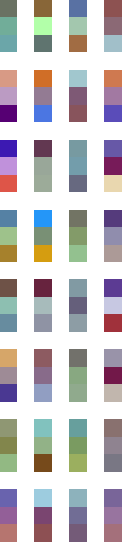}
}&
\subcaptionbox{Learned linear combinations.\label{fig:linearcomb}}
{
   \includegraphics[width=0.3\columnwidth]{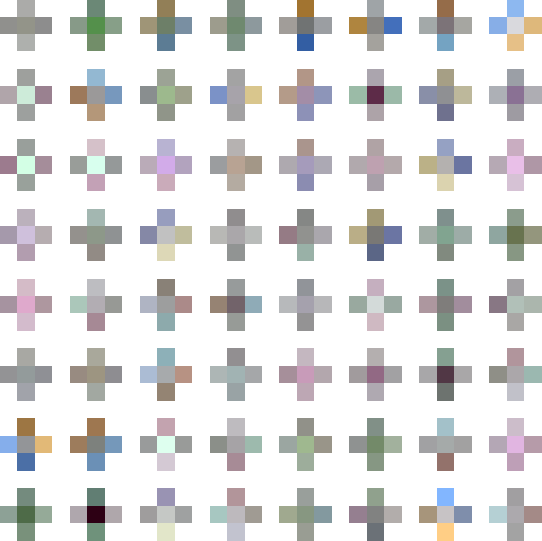}
}\\
\end{tabular}
\caption[Learned Cross-Shaped Filters.]{{\bf Learned Cross-Shaped Filters}. The cross-shaped filters (c) learned as weighted linear combination of (b) $1 \times 3$ and (c) $3 \times 1$ basis filters in the first convolutional layer of the the `vgg-gmp-lr-join' model trained using the ILSVRC dataset.}
\label{fig:conv1filters}
\end{figure}

This is illustrated in Fig.~\ref{fig:ourmethod}. Here, our composite layer contains horizontal $w\times 1$ and vertical $1\times h$ filters, the outputs of which are concatenated in the channel dimension, resulting in an intermediate $m$-channel feature map. These filter responses are then linearly combined by the next layer of $d$ $1\times 1$ filters to give a $d$-channel output feature map. In this case, the filters are applied on the input feature map with $c$ channels and followed by a set of $m$ 1$\times$1 filters over the $m$ output channels of the basis filters. If the number of horizontal and vertical filters is the same, the computational complexity is $\mathcal{O}( m(wc/2 +hc/2 + d))$.
Interestingly, the configuration of Fig.~\ref{fig:ourmethod} gives rise to linear combinations of horizontal and vertical filters that are cross-shaped in the spatial domain. This is illustrated in Fig.~\ref{fig:conv1filters} for filters learned in the first convolutional layer of the`vgg-gmp-lr-join' model that is described in the Results section when it is trained using ILSVRC dataset.

Note that, in general, more than two different sizes of basis filter might be used in the composite layer. For example,  Fig.~\ref{fig:ourmethodfullrank} shows a combination of three sets of filters with spatial dimensions $w \times 1$, $1 \times h$, and $w \times h$. Also note that an interesting option is to omit the $1 \times 1$ linear combination layer and instead allow the connection weights in a subsequent network layer to learn to combine the basis filters of the preceding layer (despite any intermediate non-linearity, \eg ReLUs). This possibility is explored in practice in the Results section.

In that our method uses a combination of filters in a composite layer, it is similar to the `GoogLeNet' of \citet{journals/corr/SzegedyLJSRAEVR14} which uses `inception' modules comprising several (square) filters of different sizes ranging from 1$\times$1 to 5$\times$5. In our case, however, we are implicitly learning linear combinations of less computationally expensive filters with different orientations (\eg 3$\times$1 and 1$\times$3 filters), rather than combinations of filters of different sizes. Amongst networks with similar computational requirements, GoogLeNet is one of the most accurate for large scale image classification tasks (see Fig.~\ref{fig:vggplots}), partly due to the use of heterogeneous filters in the inception modules, but also the use of low-dimensional embeddings and global pooling.

%% file: training.tex
\section{Training CNNs with mixed-shape low-rank filters}
\label{initialization}
To determine the standard deviations to be used for weight initialization, we use an approach similar to that described by \citet{glorot2010understanding} (with the adaptation described by \citet{journals/corr/HeZR015} for layers followed by a ReLU). In Appendix \ref{initializationderivation}, we show the details of our derivation, generalizing the approach of \citet{journals/corr/HeZR015} to the initialization of `composite' layers comprising several groups of filters of different spatial dimensions (see Appendix \ref{initializationderivation}, Fig.~\ref{fig:compositelayers}). This is one of the main contributions of this work.

We find that a composite layer of heterogeneously-shaped filter groups, where each filter group $i$ has $w^{[i]} h^{[i]} d^{[i]}$ outgoing connections should be initialized as if it is a single layer with  $\hat{n} = \sum{ w^{[i]} h^{[i]} d^{[i]}}$. Thus in the case of a ReLU non-linearity, we find that such a composite layer should be initialized with a zero-mean Gaussian distribution with standard deviation:

\begin{equation}
\sigma = \sqrt{\frac{2}{\sum{ w^{[i]} h^{[i]} d^{[i]}}}}.
\end{equation}

%% file: results.tex
\section{Results and Comparisons}
To validate our approach, we show that we can replace the filters used in existing state-of-the-art network architectures with low-rank representations as described above to reduce computational complexity without reducing accuracy. Here we characterize the computational complexity of a CNN using the number of multiply accumulate operations required for a forward pass (which depends on the size of the filters in each convolutional layer as well as the input image size and stride). However, we have observed strong correlation between multiply-accumulate counts and runtime for both CPU and GPU implementations of the networks described here (as shown in Appendix \ref{mavstimings}, Fig.~\ref{fig:mavstimings}). Note that the Caffe timings differ more for the initial convolutional layers where the input sizes are much smaller (3-channels) as \text{BLAS} is less efficient for the relatively small matrices being multiplied.

\paragraph{Methodology.} We augment our training set with randomly cropped and mirrored images, but do not use any scale or photometric augmentation, or over-sampling. This allows us to compare the efficiency of different network architectures without having to factor in the computational cost of the various augmentation methods used elsewhere. During training, for every model except GoogLeNet, we adjust the learning rate according to the schedule $\gamma_t = \gamma_0(1+\gamma_0\lambda t)^{-1}$, where $\gamma_0,\gamma_t$ and $\lambda$ are the initial learning rate, learning rate at iteration $t$, and weight decay respectively~\citep{bottou2012stochastic}. When the validation accuracy levels off we manually reduce the learning rate by further factors of 10 until the validation accuracy no longer increases. Unless otherwise indicated, aside from changing the standard deviation of the normally distributed weight initialization, as explained in \S\ref{initialization}, we used the standard hyper-parameters for each given model. Our results use no test-time augmentation.  

\subsection{VGG-11 Architectures for ILSVRC Object Classification and MIT Places Scene Classification}
\label{vggresults}
We evaluated classification accuracy of the VGG-11 based architectures using two datasets, ImageNet Large Scale Visual Recognition Challenge 2012 (`ILSVRC') and MIT Places. The ILSVRC dataset comprises 1.2M training images of 1000 object classes, commonly evaluated by top-1 and top-5 accuracy on the 50K image validation set. The MIT Places dataset comprises 2.4M training images from 205 scene classes, evaluated with top-1 and top-5 accuracy on the 20K image validation set.

VGG-11 (`VGG-A') is an 11-layer convolutional network introduced by \citet{journals/corr/SimonyanZ14a}. It is in the same family of network architectures used by \citet{journals/corr/SimonyanZ14a,journals/corr/HeZR015} to obtain the state-of-the-art accuracy for ILSVRC, but uses fewer convolutional layers and therefore fits on a single GPU during training. During training of our VGG-11 based models, we used the standard hyperparameters as detailed by \citet{journals/corr/SimonyanZ14a} and the initialization of \citet{journals/corr/HeZR015}.

In what follows, we compare the accuracy of a number of different network architectures detailed in Appendix~\ref{vggmodeltable}, Table~\ref{table:vggarch}. Results for ILSVRC are given in Table~\ref{table:vggimagenetresults}, and plotted in Fig.~\ref{fig:vggplots}. Results for MIT Places are given in Table \ref{table:placesresults}, and plotted in Fig.~\ref{fig:placesresults}. 

\paragraph{Baseline (Global Max Pooling).}  Compared to the version of the network described in \citep{journals/corr/SimonyanZ14a}, we use a variant that replaces the final $2 \times 2$ max pooling layer before the first fully connected layer with a global max pooling operation, similar to the global average pooling used by \citet{journals/corr/LinCY13,journals/corr/SzegedyLJSRAEVR14}. We evaluated the accuracy of the baseline VGG-11 network with global max-pooling (\textbf{vgg-gmp}) and without (\textbf{vgg-11}) on the two datasets. We trained these networks at stride 1 on the ILSVRC dataset and at stride 2 on the larger MIT Places dataset. This globally max-pooled variant of VGG-11 uses over 75\% fewer parameters than the original network and gives consistently better accuracy -- almost 3 percentage points lower top-5 error on ILSVRC than the baseline VGG-11 network on ILSVRC (see Table~\ref{table:vggimagenetresults}). We used this network as the baseline for the rest of our experiments.

\paragraph{Separable Filters.} To evaluate the separable filter approach described in \S \ref{seqsep} (illustrated in Fig.~\ref{fig:separablemethods}b), we replaced each convolutional layer in VGG-11 with a sequence of two layers, the first containing horizontally oriented $1 \times 3$ filters and the second containing vertically oriented $3 \times 1$ filters (\textbf{vgg-gmp-sf}). These filters applied in sequence represent $3 \times 3$ kernels using a low dimensional basis space. Unlike \citet{journals/corr/JaderbergVZ14}, we trained this network from scratch instead of approximating the full-rank filters in a previously trained network. Compared to the original VGG-11 network, the separable filter version requires approximately 14\% less compute. Results are shown in Table~\ref{table:vggimagenetresults} for ILSVRC and Table~\ref{table:placesresults} for MIT Places. Accuracy for this network is approx.~0.8\% lower than that of the baseline vgg-11-gmp network for ILSVRC and broadly comparable for MIT Places. This approach does not give such a significant reduction in computational complexity as what follows, but it is nonetheless interesting that separable  filters are capable of achieving quite high classification accuracy on such challenging tasks.

\paragraph{Simple Horizontal/Vertical Basis.} To demonstrate the efficacy of the simple low rank filter representation illustrated in Fig.~\ref{fig:separablemethods}c, we created a new network architecture (\textbf{vgg-gmp-lr-join}) by replacing each of the convolutional layers in VGG-11 (original filter dimensions were $3 \times 3$) with a sequence of two layers. The first layer comprises half $1\times 3$ filters and half $3\times 1$ filters whilst the second layer comprises the same number of $1\times 1$ filters. The resulting network is approximately 49\% faster than the original and yet it gives broadly comparable accuracy (within 1 percentage point) for both the ILSVRC and MIT Places datasets.

\paragraph{Full-Rank Mixture.} An interesting question concerns the impact on accuracy of combining a small proportion of 3×3 filters with the 1×3 and 3×1 ﬁlters used in ‘vgg-gmp-lr-join’. To answer this question, we trained a network, \textbf{vgg-gmp-lr-join-wfull}, with a mixture of 25\% $3 \times 3$ and 75\% $1 \times 3$ and $3 \times 1$ filters, while preserving the total number of filters of the baseline network (as illustrated in Fig.~\ref{fig:ourmethodfullrank}). This network was significantly more accurate than both `vgg-gmp-lr-join' and the baseline, with a top-5 center crop accuracy of 89.7\% on ILSVRC, with a computational savings of approx.~16\% over our baseline. We note that the accuracy is approx.~1 percentage point higher than GoogLeNet.

\paragraph{Implicitly Learned Combinations.} In addition, we try a network similar to vgg-gmp-lr-join but without the $1 \times 1$ convolutional layer (as shown in Fig.~\ref{fig:ourmethod}) used to sum the contributions of $3 \times 1$ and $1 \times 3$ filters (\textbf{vgg-gmp-lr}). Interestingly, because of the elimination of the extra $1\times 1$ layers, this gives an additional compute saving such that this model is is only $1/3^{\textrm{rd}}$ of the compute of our baseline, with no reduction in accuracy. This seems to be a consequence of the fact that the subsequent convolutional layer is itself capable of learning effective combinations of filter responses even after the intermediate ReLU non-linearity.

We also trained such a network with double the number of convolutional filters (\textbf{vgg-gmp-lr-2x}), \ie with an equal number of $1 \times 3$ and $3 \times 1$ filters, or $2c$ filters as shown in Fig.~\ref{fig:ourmethod}. We found this to increase accuracy further (88.9\% Top-5 on ILSVRC) while still being approximately 58\% faster than our baseline network.

\paragraph{Low-Dimensional Embeddings.}
We attempted to reduce the computational complexity of our `gmp-lr' network further in the \textbf{vgg-gmp-lr-lde} network by using a stride of 2 in the first convolutional layer, and adding low-dimensional embeddings, as in \citet{journals/corr/LinCY13,journals/corr/SzegedyLJSRAEVR14}. We reduced the number of output channels by half after each convolutional layer using $1 \times 1$ convolutional layers, as detailed in Appendix~\ref{vggmodeltable}, Table~\ref{table:vggarch}. While this reduces computation significantly, by approx.~86\% compared to our baseline, we saw a decrease in top-5 accuracy on ILSVRC of 1.2 percentage points. We do note however, that this network remains 2.5 percentage points more accurate than the original VGG-11 network, but is 87\% faster.

\input{vggresults}
\input{vggplots}

\input{vggplacesresults}

\subsection{GoogLeNet for ILSVRC Object Classification}
GoogLeNet, introduced by \citet{journals/corr/SzegedyLJSRAEVR14}, is the most efficient network for ILSVRC, getting close to state-of-the-art results with a fraction of the compute and model size of even VGG-11. The GoogLeNet inception module is a composite layer of 5 homogeneously-shaped filters, $1\times 1$, $3\times 3$, $5\times 5$, and the output of a 3x3 average pooling operations. All of these are concatenated and used as input for successive layers. 

For the \textbf{googlenet-lr} network, within only the inception modules we replaced each the $3\times 3$ filters with low-rank $3 \times 1$ and $1\times 3$ filters, and replaced the layer of $5\times 5$ filters with a set of low-rank $5 \times 1$ and $1\times 5$ filters. For the \textbf{googlenet-lr-conv1} network, we similarly replaced the first and second layer convolutional layers with $7 \times 1$~/~$1\times 7$ and $3 \times 1$~/~$1\times 3$ layers respectively.

Results are shown in Table~\ref{table:googlenetimagenetresults}. Due to the intermediate losses used for training, which contain the only fully-connected layers in GoogLeNet, test time model size is significantly smaller than training time model size. Table~\ref{table:googlenetimagenetresults} also reports test time model size. The low-rank network delivers comparable classification accuracy using 26\% less compute.  No other networks produce comparable accuracy within an order of magnitude of compute. We note that although the Caffe pre-trained GoogLeNet model~\citep{jia2014caffe} has a top-5 accuracy of 0.889, our training of the same network using the given model definition, including the hyper-parameters and training schedule, but a different random initialization had a top-5 accuracy of 0.883.

\input{googlenetresults}

\subsection{Network-in-Network for CIFAR-10 Object Classification}
The CIFAR-10 dataset consists of 60,000 $32\times 32$ images in 10 classes, with 6000 images per class. This is split into standard sets of 50,000 training images, and 10,000 test images~\citep{krizhevsky2009learning}. As a baseline for the CIFAR-10 dataset, we used the Network in Network architecture~\citep{journals/corr/LinCY13}, which has a published test-set error of 8.81\%. We also used random crops during training, with which the network has an error of 8.1\%. Like most state of the art CIFAR results, this was with ZCA pre-processed training and test data~\citep{goodfellow2013maxout}, training time mirror augmentation and random sub-crops. The results of our CIFAR experiments are listed in Table~\ref{table:cifarresults} and plotted in Fig.~\ref{fig:cifarresults}.

\input{cifarresults}

This architecture uses $5\times5$ filters in some layers. We found that we could replace all of these with $3\times 3$ filters, with comparable accuracy. As suggested by \citet{journals/corr/SimonyanZ14a}, stacked  $3\times 3$ filters have the effective receptive field of larger filters with less computational complexity. In this \textbf{nin-c3} network, we replaced the first convolutional layer with one $3\times 3$ layer, and the second convolutional layer with two $3\times 3$ layers. This network is 26\% faster than the standard NiN model, with only 54\% of the model parameters. Using our low-rank filters in this network, we trained the \textbf{nin-c3-lr} network, which is of similar accuracy (91.8\% \vs 91.9\%) but is approximately 54\% of the original network's computational complexity, with only 45\% of the model parameters.

%% file: vggresults.tex
\begin{table}[htb]
\centering
\pgfplotstableread[col sep=comma]{data/vggma.csv}\data
\pgfplotstabletypeset[
    every head row/.style={
    before row=\toprule,after row=\midrule},
    every last row/.style={
    after row=\bottomrule},
    every first row/.style={
    after row=\bottomrule},
    fixed zerofill,     
    columns={Network, Stride, Multiply-Acc., Param., Top-1 Acc., Top-5 Acc.},
    column type/.add={lrrrrrr}{},
    columns/Multiply-Acc./.style={
        column name=Multiple-Acc. {\small $\times 10^{9}$},
        preproc/expr={{##1/1e9}}
    },
    columns/Param./.style={
        column name=Param. {\small $\times 10^{7}$},
        preproc/expr={{##1/1e7}}
    },
    columns/Network/.style={string type},
    columns/Stride/.style={precision=0},
    columns/Top-1 Acc./.style={precision=3},
    columns/Top-5 Acc./.style={precision=3},
    highlight col max ={\data}{Top-1 Acc.},
    highlight col max ={\data}{Top-5 Acc.}, 
    highlight col min ={\data}{Param.}, 
    highlight col min ={\data}{Multiply-Acc.}, 
    col sep=comma]{\data}
\caption[VGG ILSVRC Results]{{\bf VGG ILSVRC Results.} Accuracy, multiply-accumulate count, and number of parameters for the baseline VGG-11 network (both with and without global max pooling) and more efficient versions created by the methods described in this paper.}
\label{table:vggimagenetresults}
\end{table}

%% file: vggplots.tex
\begin{figure}[htb!] 
\centering
\pgfplotstableread[col sep=comma]{data/vggma.csv}\datatable
\pgfplotsset{major grid style={dotted,red}}

\begin{tikzpicture}
\begin{axis}[
  width=\columnwidth,
  height=0.33\columnwidth,
  axis x line=bottom,
  ylabel=Top-5 Error,
  xlabel=Multiply-Accumulate Operations,
  axis lines=left,
  enlarge x limits=0.10,
  grid=major,
  ytick={0.01,0.02,...,0.21},
  ymin=0.1,ymax=0.15,
  yticklabel={\pgfmathparse{\tick*100}\pgfmathprintnumber{\pgfmathresult}\%},style={
        /pgf/number format/fixed,
        /pgf/number format/precision=1
  },
  legend style={at={(0.01,0.98)}, anchor=north west, column sep=0.5em},
  legend columns=2,
]
\addplot[mark=square*,mark options={fill=blue},nodes near coords,only marks,
   point meta=explicit symbolic,
   x filter/.code={
       \ifnum\coordindex>2\def\pgfmathresult{}\fi
   },
] table[meta=Network,x=Multiply-Acc.,y expr={1 - \thisrow{Top-5 Acc.} },]{\datatable};
\addplot[mark=*,mark options={fill=green},nodes near coords,only marks,
   point meta=explicit symbolic,
   x filter/.code={
       \ifnum\coordindex<3\def\pgfmathresult{}\fi
   },
] table[meta=Network,x=Multiply-Acc.,y expr={1 - \thisrow{Top-5 Acc.} },]{\datatable};
\legend{Baseline Networks, Our Results}
\end{axis}
\end{tikzpicture}
\caption{\textbf{VGG ILSVRC Results.} Multiply-accumulate operations \vs top-5 error for VGG-derived models on ILSVRC object classification dataset, the most efficient networks are closer to the origin. Our models are significantly faster than the baseline network, in the case of `gmp-lr-2x' by a factor of almost 60\%, while slightly lowering error. Note that the `gmp-lr' and `gmp-lr-join' networks have the same accuracy, showing that an explicit linear combination layer may be unnecessary.}
\label{fig:vggplots}
\end{figure}
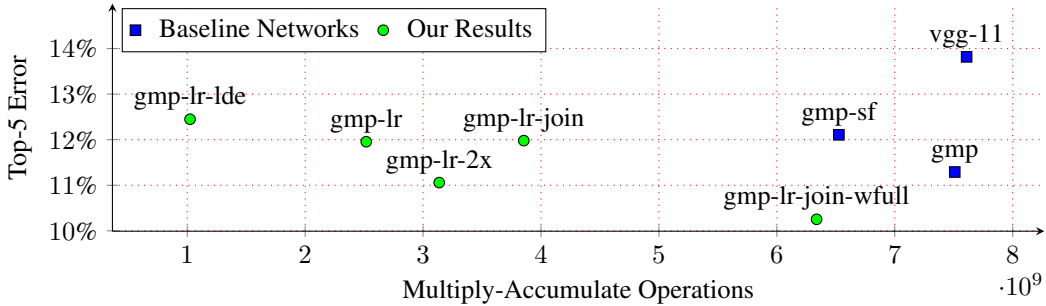

%% file: vggplacesresults.tex
\begin{table}[htbp]
\centering
\pgfplotstableread[col sep=comma]{data/mitma.csv}\data
\pgfplotstabletypeset[
    every head row/.style={
    before row=\toprule,after row=\midrule},
    every last row/.style={
    after row=\bottomrule},
    fixed zerofill,     
    columns={Network, Stride, Multiply-Acc., Param., Top-1 Acc., Top-5 Acc.},
    column type/.add={lrrrrrr}{},
    columns/Multiply-Acc./.style={
        column name=Multiple-Acc. {\small $\times 10^{8}$},
        preproc/expr={{##1/1e8}}
    },
    columns/Param./.style={
        column name=Param. {\small $\times 10^{7}$},
        preproc/expr={{##1/1e7}}
    },
    columns/Network/.style={string type},
    columns/Stride/.style={precision=0},
    columns/Top-1 Acc./.style={precision=3},
    columns/Top-5 Acc./.style={precision=3},
    highlight col max ={\data}{Top-1 Acc.},
    highlight col max ={\data}{Top-5 Acc.}, 
    highlight col min ={\data}{Param.}, 
    highlight col min ={\data}{Multiply-Acc.}, 
    col sep=comma]{\data}
\caption[MIT Places results]{{\bf MIT Places Results.} Accuracy, multiply-accumulate operations, and number of parameters for the baseline `vgg-11-gmp' network, separable filter network as described by \citet{journals/corr/JaderbergVZ14}, and more efficient models created by the methods described in this paper. All networks were trained at stride 2 for the MIT Places dataset.
 }
\label{table:placesresults}
\end{table}

%% file: googlenetresults.tex
\begin{table}[htb]
\centering
\pgfplotstableread[col sep=comma]{data/googlenetma.csv}\data
\pgfplotstabletypeset[
    every head row/.style={
    before row=\toprule,after row=\midrule},
    every last row/.style={
    after row=\bottomrule},
    every first row/.style={
    after row=\bottomrule}, 
    fixed zerofill,     
    columns={Network, Multiply-Acc., Test Param., Top-1 Acc., Top-5 Acc.},
    columns/Multiply-Acc./.style={
        column name=Multiple-Acc. {\small $\times 10^{9}$},
        preproc/expr={{##1/1e9}}
    },
    columns/Test Param./.style={
        column name=Test Param. {\small $\times 10^{6}$},
        preproc/expr={{##1/1e6}}
    },
    column type/.add={lrrrrrr}{},
    columns/Network/.style={string type},
    columns/Top-1 Acc./.style={precision=3},
    columns/Top-5 Acc./.style={precision=3},
    highlight col max ={\data}{Top-1 Acc.},
    highlight col max ={\data}{Top-5 Acc.}, 
    highlight col min ={\data}{Test Param.}, 
    highlight col min ={\data}{Multiply-Acc.}, 
    col sep=comma]{\data}
\caption[GoogLeNet ILSVRC Results]{{\bf GoogLeNet ILSVRC Results.} Accuracy, multiply-accumulate count, and number of parameters for the baseline GoogLeNet network and more efficient versions created by the methods described in this paper.
}
\label{table:googlenetimagenetresults}
\end{table}

%% file: cifarresults.tex
\begin{table}[htb]
\centering
\pgfplotstableread[col sep=comma]{data/cifarma.csv}\data
\pgfplotstabletypeset[
    every head row/.style={
    before row=\toprule,after row=\midrule},
    every last row/.style={
    after row=\bottomrule},    
    every first row/.style={
    after row=\bottomrule},
    fixed zerofill,     
    columns={Network, Multiply-Acc., Param., Accuracy},
    columns/Multiply-Acc./.style={
        column name=Multiple-Acc. {\small $\times 10^{8}$},
        preproc/expr={{##1/1e8}}
    },
    columns/Param./.style={
        column name=Param. {\small $\times 10^{5}$},
        preproc/expr={{##1/1e5}}
    },
    column type/.add={lrrr}{},
    columns/Network/.style={string type},
    columns/Accuracy/.style={precision=4}, 
    highlight col max ={\data}{Accuracy}, 
    highlight col min ={\data}{Param.}, 
    highlight col min ={\data}{Multiply-Acc.}, 
    col sep=comma]{\data}
\caption[CIFAR-10 Results.]{{\bf Network-in-Network CIFAR-10 Results.} Accuracy, multiply-accumulate operations, and number of parameters for the baseline Network-in-Network model and more efficient versions created by the methods described in this paper.}
\label{table:cifarresults}
\end{table}

%% file: conclusion.tex
\section{Discussion}

It is somewhat surprising that networks based on learning filters with less representational ability are able to do as well, or better, than CNNs with full $k\times k$ filters on the task of image classification. However, a lot of interesting small-scale image structure is well-characterized by low-rank filters, \eg edges and gradients. Our experiments training a separable (rank-1) model (`vgg-gmp-sf') on ILSVRC and MIT Places show surprisingly high accuracy on what are considered challenging problems -- approx.\ 88\% top-5 accuracy on ILSVRC -- but not enough to obtain comparable accuracies to the models on which they are based.

Given that most discriminative filters learned for image classification appear to be low-rank, we instead structure our architectures with a set of basis filters in the way illustrated in Fig.~\ref{fig:ourmethodfullrank}. This allows our networks to learn the most effective combinations of complex (\eg $k\times k$) and simple (\eg $1\times k$, $k\times 1$) filters. Furthermore, in restricting how many complex spatial filters may be learned, this architecture prevents over-fitting, and helps improve generalization. Even in our models where we do not use square $k\times k$ filters, we obtain comparable accuracies to the baseline model, since the rank-2 cross-shaped filters effectively learned as a combination of $3 \times 1$ and $1 \times 3$ filters are capable of representing more complex local pixel relations than rank-1 filters.

\section{Conclusion}
This paper has presented a method to train convolutional neural networks from scratch using low-rank filters. This is made possible by a new way of initializing the network’s weights which takes into consideration the presence of differently shape filters in composite layers. 
Validation on image classification in three popular datasets confirms similar or higher accuracy than state of the art, with much greater computational efficiency. 

Recent advances in state-of-the-art accuracy with CNNs for image classification have come at the cost of increasingly large and computational complex models. We believe our results to show that learning computationally efficient models with fewer, more relevant parameters, can prevent over-fitting, increase generalization and thus also increase accuracy.

\subsection*{Future Work}
This paper has addressed the spatial extents of convolutional filters in CNNs, however the channel extents also exhibit some redundancy, as highlighted by \citet{journals/corr/JaderbergVZ14}, and exploited in the form of low-dimensional embeddings by \citet{journals/corr/LinCY13,journals/corr/SzegedyLJSRAEVR14}. We intend to further explore how our methods can be extended to learn and combine even smaller basis filters and filters with more diverse shapes.

%% file: initialization.tex
\section{Initializing CNNs with mixed-shape low-rank filters}
\label{initializationderivation}
At the start of training, network weights are initialized at random using samples drawn from a Gaussian distribution with a standard deviation parameter specified separately for each layer. We found that the setting of these parameters was critical to the success of network training and difficult to get right, particularly because published parameter settings used elsewhere were not suitable for our new network architectures. With unsuitable weight initialization, training may fail due to {\em exploding gradients}, where  back propagated gradients grow so large as to cause numeric overflow, or {\em vanishing gradients} where back propagated gradients grow so small that their effect is dwarfed by that of weight decay such that loss does not decrease during training \citep{Hochreiter01gradientflow}.

To determine the standard deviations to be used for weight initialization, we use an approach similar to that described by \citet{glorot2010understanding} (with the adaptation described by \citet{journals/corr/HeZR015} for layers followed by a ReLU). Their approach works by ensuring that the magnitudes of back-propagated gradients remain approximately the same throughout the network. Otherwise, if the gradients were inappropriately scaled by some factor (\eg $\beta$) then the final back-propagated signal would be scaled by a potentially much larger factor ($\beta^L$ after $L$ layers).

\begin{figure}[tb!]
\centering
\includegraphics[width=0.7\columnwidth]{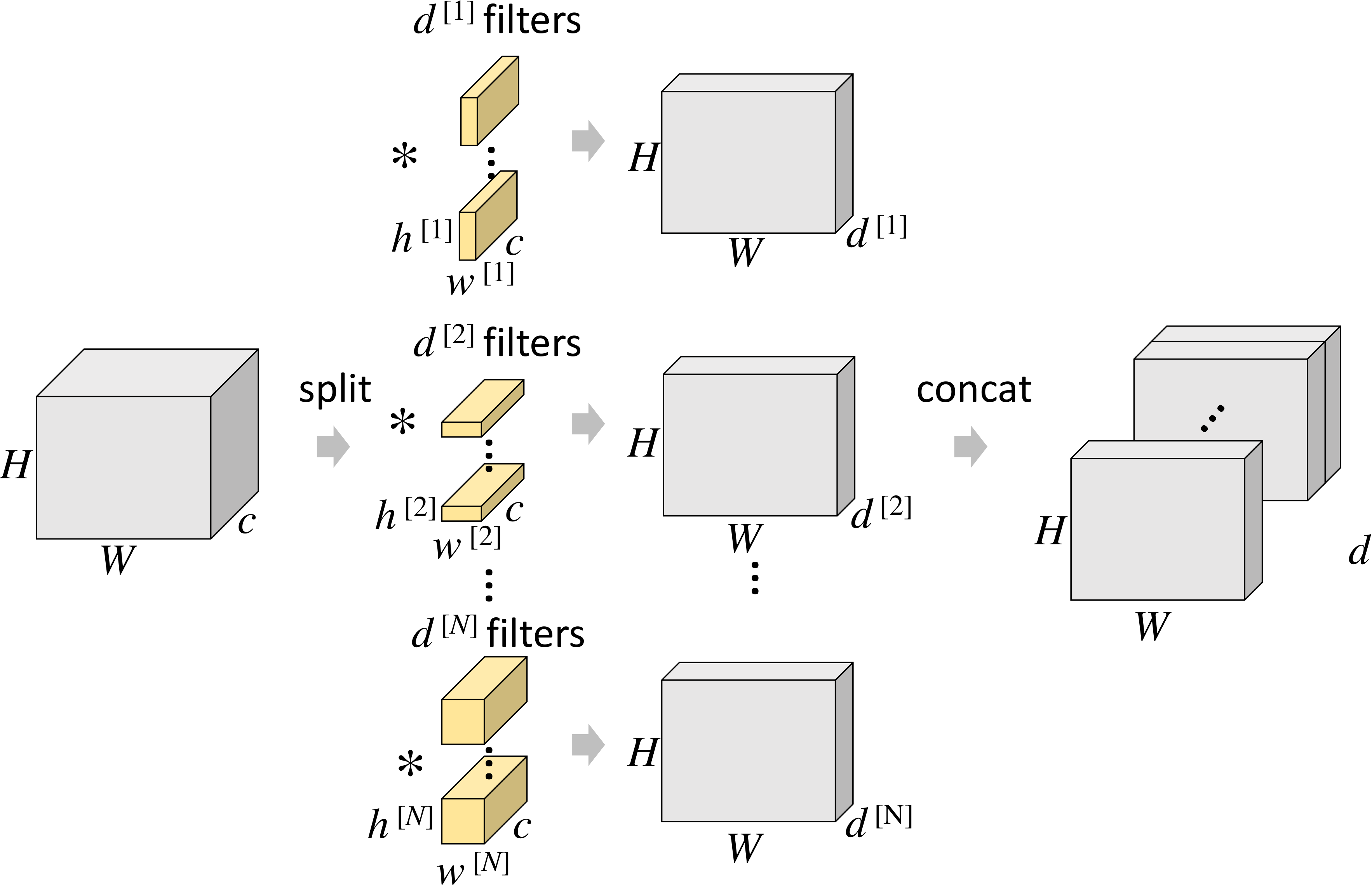}
\caption[Composite Layer.]{{\bf A composite layer}. Composite layers convolve an input feature map with $N$ groups of convolutional filters of several different spatial dimensions. Here the $i^\text{th}$ group has $d^{[i]}$ filters with spatial dimension $w^{[i]} \times h^{[i]}$. The outputs are concatenated to create a $d$ channel output feature map. Composite layers require careful weight initialization to avoid vanishing/exploding gradients during training.}
\label{fig:compositelayers}
\end{figure}

In what follows, we adopt notation similar to that of \citet{journals/corr/HeZR015}, and follow their derivation of the appropriate standard deviation for weight initialization. However, we also generalize their approach to the initialization of `composite' layers comprising several groups of filters of different spatial dimensions (see Fig.~\ref{fig:compositelayers}). This is one of the main contributions of this work.

\paragraph{Forward Propagation.}
 The response of the $l^\text{th}$ convolutional layer can be represented as
\begin{equation}
	\mathbf{y}_l =\mathbf{W}_l \mathbf{x}_l + \mathbf{b}_l.
\end{equation}
Here $\mathbf{y}_l$ is a $d \times 1$ vector representing a pixel in the output feature map and $\bf x_l$ is a $ w h c \times 1$ vector that represents a $w \times h$ subregion of the $c$-channel input feature map. $\mathbf{W}_l$ is the $d\times n$ weight matrix, where $d$ is the number of filters and $n$ is the size of a filter, \ie $n = w h c$ for a filter with spatial dimensions $w \times h$ operating on an input feature map of $c$ channels, and $\mathbf{b}_l$ is the bias. Finally $\mathbf{x}_l = f(\mathbf{y}_{l-1})$ is the output of the previous layer passed through an activation function $f$ (\eg the application of a ReLU to each element of $\mathbf{y}_{l-1}$).

\paragraph{Backward Propagation.}
During back-propagation, the gradient of a convolutional layer is computed as
\begin{equation}
\Delta \mathbf{x}_l = \hat{\mathbf{W}}_l \Delta \mathbf{y}_l,
\label{eq:back_prop_gradient}
\end{equation}
where $\Delta \mathbf{x}_l$ and $\Delta \mathbf{y}_l$ denote the derivatives of loss $\cal L$ with respect to input and output pixels. $\Delta \mathbf{x}_l$ is a $c \times 1$ vector of gradients with respect to the channels of a single pixel in the input feature map and $\Delta \mathbf{y}$ represents $h \times w$ pixels in d channels of the output feature map. $\hat{\mathbf{W}}_l$ is a $c \times \hat{n}$ matrix where the filter weights are arranged in the right order for back-propagation, and $\hat{n} = whd$. Note that $\hat{\mathbf{W}}_l$ can be simply reshaped from $\mathbf{W}_l^\top$. Also note that the elements of $\Delta \mathbf{y}_l$ correspond to pixels in the output image that had a forwards dependency on the input image pixel corresponding to $\Delta \mathbf{x}$. In back propagation, each element $\Delta y_l$ of $\Delta \mathbf{y}_l$ is related to an element $\Delta x_{l+1}$ of some $\Delta \mathbf{x}_{l+1}$ (\ie a back-propagated gradient in the next layer) by the derivative of the activation function $f$:
\begin{equation}
\Delta y_l = f^\prime (y_l) \Delta x_{l+1},
\end{equation}
where $f^\prime$ is the derivative of the activation function.

\newcommand{\Expect}{\mathrm{E}}
\newcommand{\Var}{\mathrm{Var}}

\paragraph{Weight Initialization.}
Now let $\Delta y_l$, $\Delta x_l$ and $w_l$ be scalar random variables that describe the distribution of elements in $\Delta \mathbf{y}_l$, $\Delta \mathbf{x}_{l}$ and $\hat{\mathbf{W}}_l$ respectively. Then, assuming $f^\prime (y_l)$ and $\Delta x_{l+1}$ are independent,
\begin{equation}
\Expect[\Delta y_l] = \Expect[f^\prime (y_l)] \Expect[ \Delta x_{l+1}].
\end{equation}

For the ReLU case, $f'(y_l)$ is zero or one with equal probability. 
Like \citet{glorot2010understanding}, we assume that $w_l$ and $\Delta y_l$ are independent of each other. Thus, equation \ref{eq:back_prop_gradient} implies that $\Delta x_l$ has zero mean for all $l$, when $w_l$ is initialized by a distribution that is symmetric around zero. Thus we have $\Expect[\Delta y_l] = \frac{1}{2}\Expect[\Delta x_{l+1}]= 0$ and also $\Expect[(\Delta y_l)^2] = \Var[\Delta y_l] = \frac{1}{2} \Var[\Delta x_{l+1}]$. Now, since each element of $\Delta \mathbf{x}_l$ is a summation of $\hat n$ products of elements of $\hat{\mathbf{W}}_l$ and elements of $\Delta \mathbf{y}_l$, we can compute the variance of the gradients in equation \ref{eq:back_prop_gradient}:

\begin{equation}
\begin{aligned}
\Var[\Delta x_l] &=  \hat{n} \Var[w_l] \Var[\Delta y_l]\\
 &= \frac{1}{2}   \hat{n} \Var[w_l] \Var [\Delta x_{l+1}].
\end{aligned}
\end{equation}

To avoid scaling the gradients in the convolutional layers (to avoid exploding or vanishing gradients), we set the ratio between these variances to 1:
\begin{equation}
\frac{1}{2} \hat{n} \Var[w_l] = 1.
\end{equation}

This leads to the result of \citet{journals/corr/HeZR015}, in that a layer with $\hat{n}_l$ connections followed by a ReLU activation function should be initialized with a zero-mean Gaussian distribution with standard deviation $\sqrt{2/ \hat{n}_l}$.

\paragraph{Weight Initialisation in Composite Layers.}

The initialization scheme described above assumes that the layer comprises filters of spatial dimension $w \times h$. Now we extend this scheme to composite convolutional layers containing $N$ groups of filters of different spatial dimensions $w^{[i]} \times h^{[i]}$ (where superscript $[i]$ denotes the group index and with $i\in \{1,\dots,N\}$). Now the layer response is the concatenation of the responses of each group of filters:
\begin{equation}
\mathbf{y}_l =\begin{bmatrix}\mathbf{W}_l^{[1]} \mathbf{x}_l^{[1]} \\ \mathbf{W}_l^{[2]} \mathbf{x}_l^{[2]} \\ \dots \\ \mathbf{W}_l^{[N]} \mathbf{x}_l^{[N]} \end{bmatrix} + \mathbf{b}_l.
\end{equation}
As before $\mathbf{y}_l$ is a $d \times 1$ vector representing the response at one pixel of the output feature map. Now each ${\bf x}^{[i]}$ is a $w^{[i]} h^{[i]} c \times 1$ vector that represents a different shaped $w^{[i]} \times h^{[i]}$ sub-region of the input feature map. Each $\mathbf{W}_l^{[i]}$ is the $c_l^{[i]}\times \hat{n}^{[i]}$ weight matrix, where $d$ is the number of filters and $\hat{n}^{[i]}$ is the size of a filter, \ie $\hat{n}^{[i]} = w^{[i]} h^{[i]} c^{[i]}$ for a filter of spatial dimension $w^{[i]} \times h^{[i]}$ operating on an input feature map of $c_l = d_{l-1}$ channels.

During back propagation, the gradient of the composite convolutional layer is computed as a summation of the contributions from each group of filters:
\begin{equation}
\Delta \mathbf{x}_l = \hat{\mathbf{W}}_l^{[1]} \Delta \mathbf{y}_l^{[1]} +  \hat{\mathbf{W}}_l^{[2]} \Delta \mathbf{y}_l^{[2]} + \dots+  \hat{\mathbf{W}}_l^{[N]} \Delta \mathbf{y}_l^{[N]},
\label{eq:back_prop_gradient_composite}
\end{equation}
where now $\Delta \mathbf{y}^{[i]}$ represents $w^{[i]} \times h^{[i]}$ pixels in $d^{[i]}$ channels of the output feature map. Each $\hat{\mathbf{W}}_l^{[i]}$ is a $c_l \times \hat{n}^{[i]}$ matrix of weights arranged appropriately for back propagation. Again, note that each $\hat{\mathbf{W}}_l^{[i]}$ can be simply reshaped from $\mathbf{W}_l^{[i]}$.

As before, each element of $\Delta \mathbf{y}_l$ is a sum over $\hat n$ products between elements of $\hat{\mathbf{W}}^{[i]}_l$ and elements of $\Delta \mathbf{y}^{[i]}_l$ and here $\hat{n}$ is given by:
\begin{equation}
\hat{n} = \sum{ w^{[i]} h^{[i]} d^{[i]}}.
\end{equation}

In the case of a ReLU non-linearity, this leads to a zero-mean Gaussian distribution with standard deviation:

\begin{equation}
\sigma = \sqrt{\frac{2}{\sum{ w^{[i]} h^{[i]} d^{[i]}}}}.
\end{equation}

In conclusion, a composite layer of heterogeneously-shaped filter groups, where each filter group $i$ has $w^{[i]} h^{[i]} d^{[i]}$ outgoing connections should be initialized as if it is a single layer with  $\hat{n} = \sum{ w^{[i]} h^{[i]} d^{[i]}}$.

%% file: mavstimings.tex
\section{Multiply-Accumulate Operations and Caffe CPU/GPU Timings.}
\label{mavstimings}
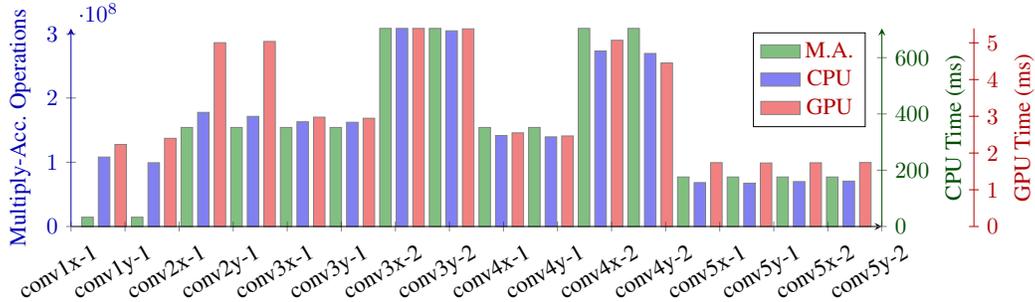
\begin{figure}[tb] 
\centering
\pgfplotstableread[col sep=comma]{data/matimings.csv}\datatable

\resizebox{\columnwidth}{!}{
\begin{tikzpicture}

\begin{axis}[
  width=\columnwidth,
  height=0.33\columnwidth,
  ybar,
  bar width=0.5em,
  axis x line=bottom,
  axis y line=none,
  xtick=data,
  xticklabels from table={\datatable}{Layer},
  x tick label style={xshift=1em, rotate=30,anchor=north east}],
  xmin=-1,xmax=48,
  xlabel=Convolutional Layer,
]
\addplot[style={mark=none,opacity=0},area legend] table[x expr=2*\coordindex,y expr={0}]{\datatable};
\end{axis}

\begin{axis}[
  width=\columnwidth,
  height=0.33\columnwidth,
  ybar,
  bar width=0.5em,
  axis x line=none,
  axis y line=left,
  xmin=-1,xmax=48,
  ymin=0,
  xlabel=Convolutional Layer,
  ylabel=Multiply-Acc.~Operations,
  ylabel near ticks,
  yticklabel pos=left,
  color=blue!70!black,
]
\addplot[style={black,fill=green!50!black,opacity=0.5,mark=none},area legend] table[x expr=3*\coordindex,y=Multiply-Accumulate]{\datatable};\label{maaxis}
\end{axis}

\begin{axis}[
  width=\columnwidth,
  height=0.33\columnwidth,
  ybar,
  bar width=0.5em,
  axis x line=none,
  axis y line=right,
  xmin=-1,xmax=48,
  ymin=0,
  ylabel=CPU Time (ms),
  ylabel near ticks,
  yticklabel pos=right,
  color=green!30!black,
]
\addplot+[style={black,fill=blue!90!black,opacity=0.5,mark=none},area legend] table[x expr=3*\coordindex+1,y=CPU Timing]{\datatable};\label{cpuaxis}
\end{axis}

\begin{axis}[
  width=\columnwidth,
  height=0.33\columnwidth,
  ybar,
  bar width=0.5em,
  axis x line=none,
  axis y line=right,
  xmin=-1,xmax=48,
  ymin=0,
  ylabel=GPU Time (ms),
  ylabel near ticks,
  yticklabel pos=right,
  color=red!70!black,
  ytick={0,1,...,5},
  legend style={at={(0.98,0.98)},anchor=north east},
]
\pgfplotsset{every outer y axis line/.style={xshift=4em}, every tick/.style={xshift=4em}, every y tick label/.style={xshift=4em} }
\addlegendimage{/pgfplots/refstyle=maaxis}\addlegendentry{M.A.}
\addlegendimage{/pgfplots/refstyle=cpuaxis}\addlegendentry{CPU}
\addplot[style={black,fill=red!90!black,opacity=0.5},area legend] table[x expr=3*\coordindex+2,y=GPU Timing]{\datatable};\label{gpuaxis}
\addlegendentry{GPU}
\end{axis}

\end{tikzpicture}
}
\caption{\textbf{Multiply-Accumulate Operations and Caffe CPU/GPU Timings.} For the forward pass of each convolutional layer in the `vgg-gmp-lr' network. Caffe CPU and GPU timings were well correlated with multiply-accumulate operations for most layers.}
\label{fig:mavstimings}
\end{figure}

We have characterized the computational complexity of a CNN using the number of multiply accumulate operations required for a forward pass (which depends on the size of the filters in each convolutional layer as well as the input image size and stride), to give as close as possible to a hardware and implementation independent evaluation the computational complexity of our method. However, we have observed strong correlation between multiply-accumulate counts and runtime for both CPU and GPU implementations of the networks described here (as shown in Fig.~\ref{fig:mavstimings}). Note that the Caffe timings differ more for the initial convolutional layers where the input sizes are much smaller (3-channels), and \text{BLAS} is less efficient for the relatively small matrices being multiplied.

%% file: bigpicture.tex
\section{Comparing with State of the Art Networks for ILSVRC}
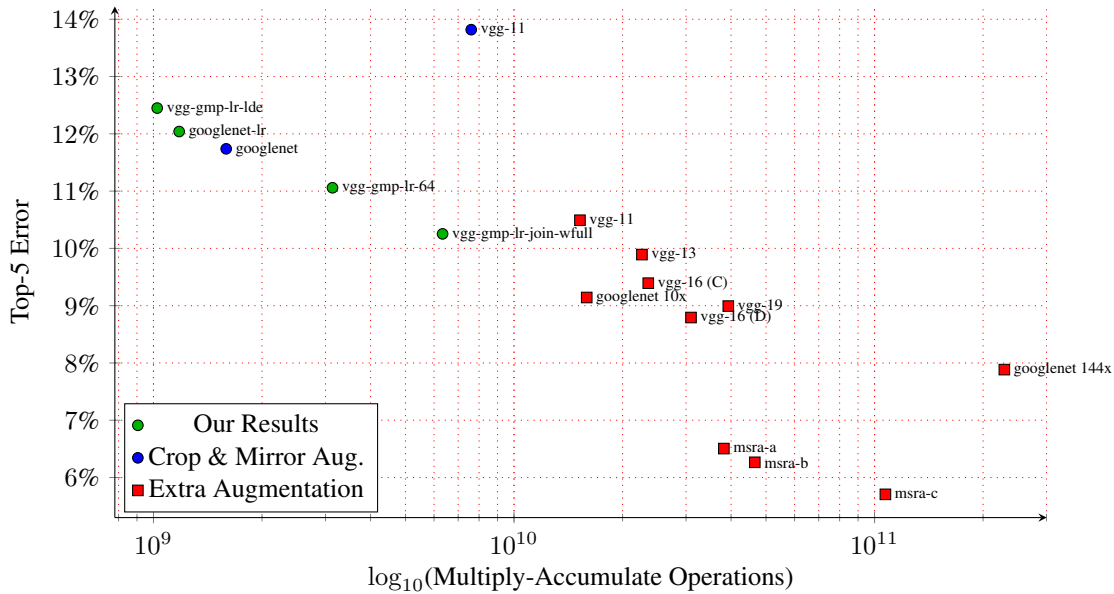
\begin{figure}[p] 
\centering
\pgfplotstableread[col sep=comma]{data/bigpicture.csv}\datatable
\pgfplotstableread[col sep=comma]{data/bigpicture_ours.csv}\datatableours
\pgfplotstableread[col sep=comma]{data/bigpicture_aug.csv}\datatableaug
\pgfplotsset{major grid style={dotted,red}}
\pgfplotsset{minor grid style={dotted,red}}

\begin{tikzpicture}
\begin{axis}[
  width=\columnwidth,
  height=0.6\columnwidth,
  axis x line=bottom,
  ylabel=Top-5 Error,
  xlabel=$\log_{10}$(Multiply-Accumulate Operations),
  axis lines=left,
  enlarge x limits=0.05,
  enlarge y limits=0.05,
  grid=both,
  ytick={0.01,0.02,...,0.2},
  xmode=log,
  yticklabel={\pgfmathparse{\tick*100}\pgfmathprintnumber{\pgfmathresult}\%},style={
        /pgf/number format/fixed,
        /pgf/number format/precision=1
  },
  legend style={at={(0.01,0.01)},anchor=south west},
]
\addplot[mark=*,mark options={fill=green!70!black},nodes near coords,only marks,
   point meta=explicit symbolic,
   every node near coord/.append style={xshift=0.01em, anchor=west, font=\tiny},
] table[meta=Network,x=Multiply-Acc.,y expr={1 - \thisrow{Top-5 Acc.} }]{\datatableours};
\addplot[mark=*,mark options={fill=blue},nodes near coords,only marks,
   point meta=explicit symbolic,
   every node near coord/.append style={xshift=0.01em, anchor=west, font=\tiny},
] table[meta=Network,x=Multiply-Acc.,y expr={1 - \thisrow{Top-5 Acc.} }]{\datatable};
\addplot[mark=square*,mark options={fill=red},nodes near coords,only marks,
   point meta=explicit symbolic,
   every node near coord/.append style={xshift=0.01em, anchor=west, font=\tiny},
] table[meta=Network,x=Test Multiply-Acc.,y expr={1 - \thisrow{Top-5 Acc.} }]{\datatableaug};
\legend{Our Results, Crop \& Mirror Aug., Extra Augmentation}
\end{axis}
\end{tikzpicture}
\caption{\textbf{Computational Complexity of Single State-of-the-Art ILSVRC Models.} Test-time multiply-accumulate operations \vs top-5 error on state of the art networks using a \emph{single} model. Note the difference in accuracy and computational complexity for VGG-11 model with/without extra augmentation. Our `vgg-gmp-lr-join-wfull' model \emph{without} extra augmentation is more accurate than VGG-11 \emph{with} extra augmentation, and is much less computationally complex.}
\label{fig:bigpicturema}
\end{figure}

\begin{figure}[p]
\centering
\pgfplotstableread[col sep=comma]{data/bigpicture.csv}\datatable
\pgfplotstableread[col sep=comma]{data/bigpicture_ours.csv}\datatableours
\pgfplotstableread[col sep=comma]{data/bigpicture_aug.csv}\datatableaug
\pgfplotsset{major grid style={dotted,red}}
\pgfplotsset{minor grid style={dotted,red}}

\begin{tikzpicture}
\begin{axis}[
  width=\columnwidth,
  height=0.6\columnwidth,
  axis x line=bottom,
  ylabel=Top-5 Error,
  xlabel=$\log_{10}$(Number of Parameters),
  axis lines=left,
  enlarge y limits=0.05,
  grid=both,
  ytick={0.01,0.02,...,0.2},
  xmode=log,
  xmin=10e5,xmax=10e8,
  yticklabel={\pgfmathparse{\tick*100}\pgfmathprintnumber{\pgfmathresult}\%},style={
        /pgf/number format/fixed,
        /pgf/number format/precision=1
  },
  legend style={at={(0.01,0.01)},anchor=south west},
]
\addplot[mark=*,mark options={fill=green!70!black},
   nodes near coords,
   only marks,
   point meta=explicit symbolic,
   every node near coord/.append style={xshift=0.01em, anchor=west, font=\tiny},
] table[meta=Network,x=Param.,y expr={1 - \thisrow{Top-5 Acc.} }]{\datatableours};
\addplot[mark=*,mark options={fill=blue},
   nodes near coords,
   only marks,
   point meta=explicit symbolic,
   every node near coord/.append style={xshift=0.01em, anchor=west, font=\tiny},
] table[meta=Network,x=Param.,y expr={1 - \thisrow{Top-5 Acc.} }]{\datatable};
\addplot[mark=square*,mark options={fill=red},
   nodes near coords,
   only marks,
   point meta=explicit symbolic,
   every node near coord/.append style={xshift=0.01em, anchor=west, font=\tiny},
   every node near coord/.append style={font=\tiny},
] table[meta=Network,x=Param.,y expr={1 - \thisrow{Top-5 Acc.} }]{\datatableaug};
\legend{Our Results, Crop \& Mirror Aug., Extra Augmentation}
\end{axis}
\end{tikzpicture}
\caption{\textbf{Number of Parameters of State-of-the-Art ILSVRC Models.} Test time parameters \vs top-5 error for state of the art models. The main factor in reduced model size is the use of global pooling or lack of fully-connected layers. Note that our `googlenet-lr' model is almost an order of magnitude smaller than any other network of comparable accuracy.}
\label{fig:bigpictureparam}
\end{figure}
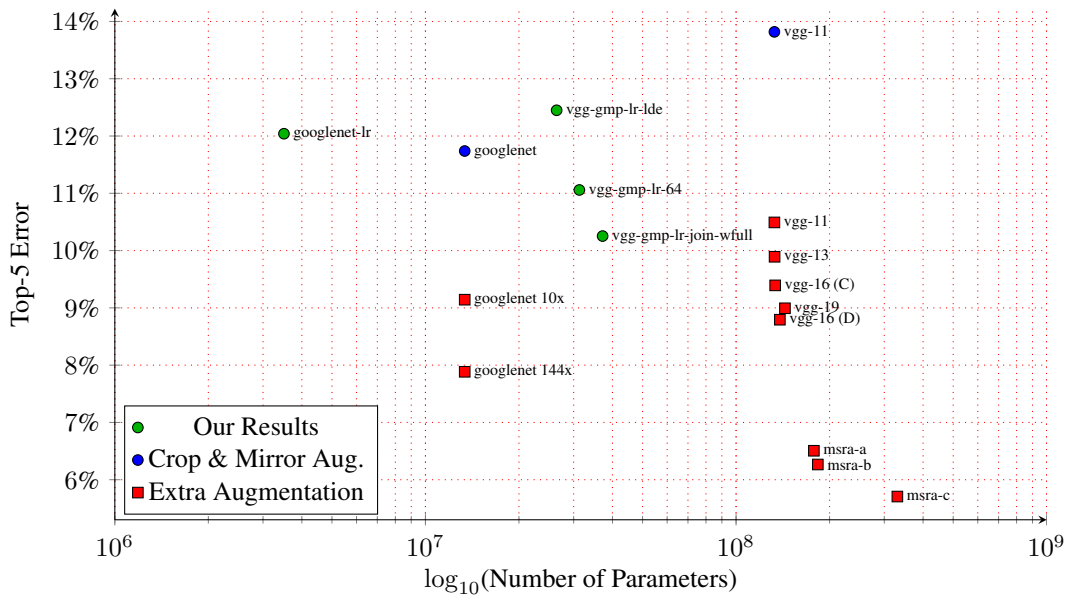

Figures ~\ref{fig:bigpicturema} and \ref{fig:bigpictureparam} compare published top-5 ILSVRC validation error \vs multiply-accumulate operations and number of model parameters (respectively) for several state-of-the-art networks~\citep{journals/corr/SimonyanZ14a,journals/corr/SzegedyLJSRAEVR14,journals/corr/HeZR015}. The error rates for these networks are only reported as obtained with different combinations of computationally expensive training and test time augmentation methods, including scale, photometric, ensembles (multi-model), and multi-view/dense oversampling. This can makes it difficult to compare model architectures, especially with respect to computational requirements.

State-of-the-art networks, such as MSRA-C, VGG-19 and oversampled GoogLeNet are orders of magnitude larger in computational complexity than our networks. From Fig.~\ref{fig:bigpicturema}, where the multiply-accumulate operations are plotted on a log scale, increasing the model size and/or computational complexity of test-time augmentation of CNNs appears to have diminishing returns for decreasing validation error. Our models \emph{without} training or test time augmentation show comparable accuracy to networks such as VGG-13 \emph{with} training and test time augmentation, while having far less computational complexity and model size. In particular, the `googlenet-lr' model has a much smaller test-time model size than any network of comparable accuracy.

\begin{table}[htb]
\centering
\pgfplotstableread[col sep=comma]{data/bigpicture_aug.csv}\data
\pgfplotstabletypeset[
    every head row/.style={
    before row=\toprule,after row=\midrule},
    every last row/.style={
    after row=\bottomrule},
    fixed zerofill,     
    columns={Network, Multiply-Acc., Test Multiply-Acc., Param., Top-5 Acc.},
    column type/.add={lrrrrrr}{},
    columns/Multiply-Acc./.style={
        column name=Multiply-Acc. {\small $\times 10^{9}$},
        preproc/expr={{##1/1e9}}
    },
    columns/Test Multiply-Acc./.style={
        column name=Test M.A. w/ Aug. {\small $\times 10^{9}$},
        preproc/expr={{##1/1e9}}
    },
    columns/Param./.style={
        column name=Param. {\small $\times 10^{7}$},
        preproc/expr={{##1/1e7}}
    },
    columns/Network/.style={string type},
    columns/Stride/.style={precision=0},
    columns/Top-1 Acc./.style={precision=3},
    columns/Top-5 Acc./.style={precision=3},
    highlight col max ={\data}{Top-5 Acc.}, 
    highlight col min ={\data}{Param.}, 
    highlight col min ={\data}{Multiply-Acc.}, 
    highlight col min ={\data}{Test Multiply-Acc.}, 
    col sep=comma]{\data}
    
\caption{{\bf State of the Art Single Models with Extra Augmentation.} Top-5 ILSVRC validation accuracy, single view and augmented test-time multiply-accumulate (M.A.) count, and number of parameters for various state of the art models \emph{with} various training and test-time augmentation methods. A multi-model ensemble of MSRA-C is the current state of the art network.}
\label{table:bigpicturetable}
\end{table}

%% file: extraplots.tex
\section{Plots of Results}
Following are several plots of results, which for reasons of space consideration are not in the main section of the  paper. These include the results for
VGG-derived models on
MIT Places (Fig.~\ref{fig:placesresults}), 
\input{vggplacesplots}
GoogLeNet-derived models on ILSVRC (Fig.~\ref{fig:googlenetimagenetresults}), 
\input{googlenetplots}
and finally the results for Network-in-Network-derived models on CIFAR-10 (Fig.~\ref{fig:cifarresults}).
\input{cifarplots}

%% file: vggplacesplots.tex
\begin{figure}[htbp] 
\centering
\pgfplotstableread[col sep=comma]{data/mitma.csv}\datatable
\pgfplotsset{major grid style={dotted,red}}

\begin{tikzpicture}
\begin{axis}[
  width=\columnwidth,
  height=0.3\columnwidth,
  axis x line=bottom,
  ylabel=Top-5 Error,
  xlabel=Multiply-Accumulate Operations,
  axis lines=left,
  grid=major,
  xmin=0.4e9,xmax=2.0e9,
  ymin=0.16,ymax=0.2,
  yticklabel={\pgfmathparse{\tick*100}\pgfmathprintnumber{\pgfmathresult}\%},style={
        /pgf/number format/fixed,
        /pgf/number format/precision=1},
  legend style={at={(0.98,0.98)}, anchor=north east, column sep=0.5em},
  legend columns=2,
]
\addplot[mark=square*,mark options={fill=blue},nodes near coords,only marks,
   point meta=explicit symbolic,
   x filter/.code={
       \ifnum\coordindex>1\def\pgfmathresult{}\fi
   }
] table[meta=Network,x=Multiply-Acc.,y expr={1 - \thisrow{Top-5 Acc.} },]{\datatable};
\addplot[mark=*,mark options={fill=green},nodes near coords,only marks,
   point meta=explicit symbolic,
   x filter/.code={
       \ifnum\coordindex<2\def\pgfmathresult{}\fi
   }
] table[meta=Network,x=Multiply-Acc.,y expr={1 - \thisrow{Top-5 Acc.} },]{\datatable};
\legend{Baseline Networks, Our Results}
\end{axis}
\end{tikzpicture}
\caption{\textbf{MIT Places Results.} Multiply-accumulate operations \vs top-5 error for VGG-derived models on MIT Places scene classification dataset. }
\label{fig:placesresults}
\end{figure}
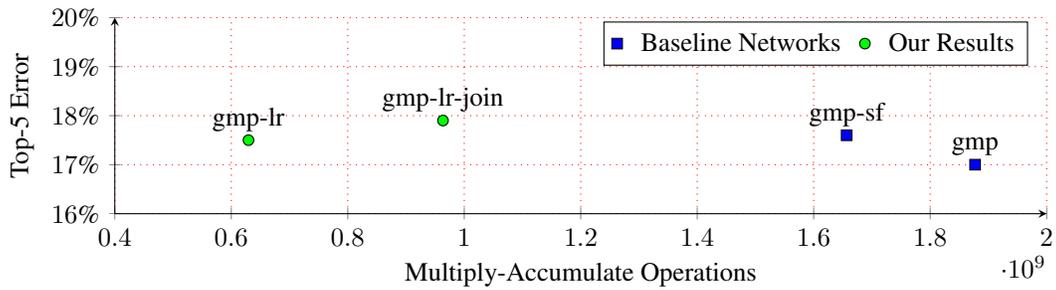

%% file: googlenetplots.tex
\begin{figure}[htbp] 
\centering
\pgfplotstableread[col sep=comma]{data/googlenetma.csv}\datatable
\pgfplotsset{major grid style={dotted,red}}

\begin{tikzpicture}
\begin{axis}[
  width=\columnwidth,
  height=0.33\columnwidth,
  axis x line=bottom,
  ylabel=Top-5 Error,
  xlabel=Multiply-Accumulate Operations,
  axis lines=left,
  enlarge x limits=0.10,
  grid=major,
  ytick={0.01,0.02,...,0.21},
  ymin=0.10,ymax=0.15,
  yticklabel={\pgfmathparse{\tick*100}\pgfmathprintnumber{\pgfmathresult}\%},style={
        /pgf/number format/fixed,
        /pgf/number format/precision=1
  },
  legend style={at={(0.98,0.98)}, anchor=north east, column sep=0.5em},
  legend columns=2,
]
\addplot[mark=square*,mark options={fill=blue},nodes near coords,only marks,
   point meta=explicit symbolic,
   x filter/.code={
       \ifnum\coordindex>0\def\pgfmathresult{}\fi
   }
] table[meta=Network,x=Multiply-Acc.,y expr={1 - \thisrow{Top-5 Acc.} },]{\datatable};
\addplot[mark=*,mark options={fill=green},nodes near coords,only marks,
   point meta=explicit symbolic,
   x filter/.code={
       \ifnum\coordindex<1\def\pgfmathresult{}\fi
   }
] table[meta=Network,x=Multiply-Acc.,y expr={1 - \thisrow{Top-5 Acc.} },]{\datatable};
\legend{Baseline, Our Results}
\end{axis}
\end{tikzpicture}

\caption{\textbf{GoogLeNet ILSVRC Results.} Multiply-accumulate operations \vs top-5 error for GoogLeNet-derived models on ILSVRC object classification dataset.}
\label{fig:googlenetimagenetresults}
\end{figure}
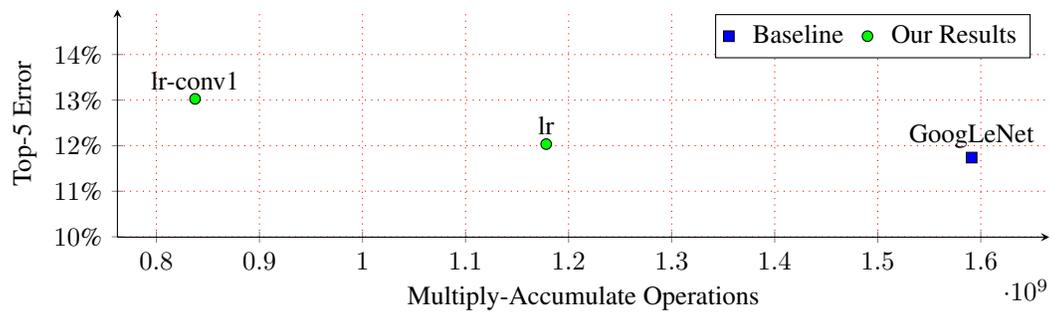

%% file: cifarplots.tex
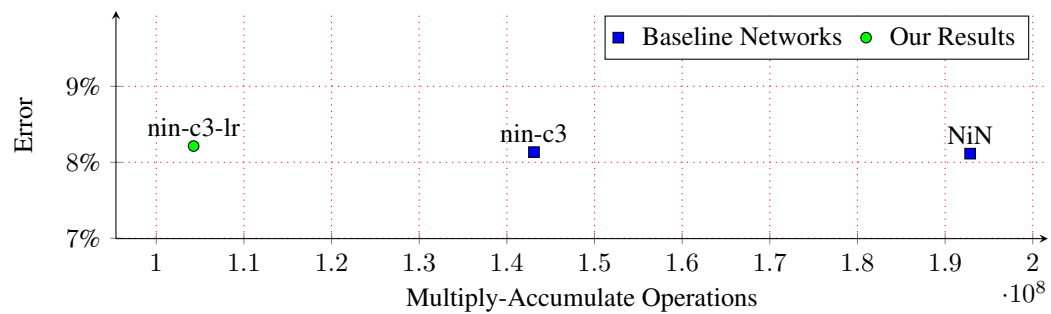
\begin{figure}[htbp] 
\centering
\pgfplotstableread[col sep=comma]{data/cifarma.csv}\datatable
\pgfplotsset{major grid style={dotted,red}}

\begin{tikzpicture}
\begin{axis}[
  width=\columnwidth,
  height=0.33\columnwidth,
  axis x line=bottom,
  ylabel=Error,
  xlabel=Multiply-Accumulate Operations,
  axis lines=left,
  enlarge x limits=0.10,
  grid=major,
  ytick={0.01,0.02,...,0.21},
  yticklabel={\pgfmathparse{\tick*100}\pgfmathprintnumber{\pgfmathresult}\%},style={
        /pgf/number format/fixed,
        /pgf/number format/precision=1
  },
  ymin=0.07,ymax=0.1,
  legend style={at={(0.98,0.98)}, anchor=north east, column sep=0.5em},
  legend columns=2,
]
\addplot[mark=square*,mark options={fill=blue},nodes near coords,only marks,
   point meta=explicit symbolic,
   x filter/.code={
       \ifnum\coordindex>1\def\pgfmathresult{}\fi
   }
] table[meta=Network,x=Multiply-Acc.,y expr={1 - \thisrow{Accuracy} }]{\datatable};
\addplot[mark=*,mark options={fill=green},nodes near coords,only marks,
   point meta=explicit symbolic,
   x filter/.code={
       \ifnum\coordindex<2\def\pgfmathresult{}\fi
   }
] table[meta=Network,x=Multiply-Acc.,y expr={1 - \thisrow{Accuracy} }]{\datatable};
\legend{Baseline Networks, Our Results}
\end{axis}
\end{tikzpicture}
\caption[Network-in-Network CIFAR-10 Results]{\textbf{Network-in-Network CIFAR-10 Results.} Multiply-accumulate operations \vs error for Network-in-Network derived models on CIFAR-10 object classification dataset.}
\label{fig:cifarresults}
\end{figure}

%% file: vggmodeltable.tex
\section{VGG-derived Model Table}
\label{vggmodeltable}
Table \ref{table:vggarch} shows the architectual details of the VGG-11 derived models used in \S\ref{vggresults}.

\begin{landscape}
{\renewcommand{\arraystretch}{1.2}
\begin{table*}[htbp!]
\centering
\resizebox{0.9\columnwidth}{!}{
\begin{tabular}{@{}|c||c|c|c|c|c|c|c|c|@{}}
	\hline
	Layer & VGG-11 & \textbf{GMP} & \textbf{GMP-SF} & \textbf{GMP-LR} & \textbf{GMP-LR-2X} & \textbf{GMP-LR-JOIN} & \textbf{GMP-LR-LDE} & \textbf{GMP-LR-JOIN-WFULL}\\
	\hline
	\hline
	\textbf{conv1}& \multicolumn{2}{c|}{3$\times$3, 64} & 1$\times$3, 64 & 3$\times$1, 32 $\|$ 1$\times$3, 32 & 3$\times$1, 64 $\|$ 1$\times$3, 64 & \multicolumn{2}{c|}{3$\times$1, 32 $\|$ 1$\times$3, 32}& 3$\times$1, 24 $\|$ 1$\times$3, 24 $\|$ 3$\times$3, 16\\
	\cline{4-4} \cline{7-9}	
    & \multicolumn{2}{c|}{} & 3$\times$1, 64 & & & 1$\times$1, 64 & 1$\times$1, 32 & 1$\times$1, 64 \\
    \cline{2-9}
	& \multicolumn{8}{c|}{ReLU}\\
    \cline{2-9}
	& \multicolumn{8}{c|}{2$\times$2 maxpool, /2}\\
	\hline
    \textbf{conv2} & \multicolumn{2}{c|}{3$\times$3, 128} & 1$\times$3, 128 & 3$\times$1, 64 $\|$ 1$\times$3, 64 & 3$\times$1, 128 $\|$ 1$\times$3, 128 & \multicolumn{2}{c|}{3$\times$1, 64 $\|$ 1$\times$3, 64} &  3$\times$1, 48 $\|$ 1$\times$3, 48 $\|$ 3$\times$3, 32\\
    \cline{4-4} \cline{7-9}
    & \multicolumn{2}{c|}{} & 3$\times$1, 128 & & & 1$\times$1, 128 & 1$\times$1, 64 & 1$\times$1, 128 \\
    \cline{2-9}
	& \multicolumn{8}{c|}{ReLU}\\
    \cline{2-9}
    & \multicolumn{8}{c|}{2$\times$2 maxpool, /2}\\
	\hline
	\textbf{conv3} & \multicolumn{2}{c|}{3$\times$3, 256} & 1$\times$3, 256 & 3$\times$1, 128 $\|$ 1$\times$3, 128 & 3$\times$1, 256 $\|$ 1$\times$3, 256 & \multicolumn{2}{c|}{3$\times$1, 128 $\|$ 1$\times$3, 128} & 3$\times$1, 96 $\|$ 1$\times$3, 96 $\|$ 3$\times$3, 64 \\
	\cline{4-4} \cline{7-9}
    & \multicolumn{2}{c|}{} & 3$\times$1, 256 & & & 1$\times$1, 256 & 1$\times$1, 128 & 1$\times$1, 256\\
    \cline{2-9}
	& \multicolumn{8}{c|}{ReLU}\\
    \cline{2-9}
	& \multicolumn{2}{c|}{3$\times$3, 256} & 1$\times$3, 256 & 3$\times$1, 128 $\|$ 1$\times$3, 128 & 3$\times$1, 256 $\|$ 1$\times$3, 256 &\multicolumn{2}{c|}{3$\times$1, 128 $\|$ 1$\times$3, 128} & 3$\times$1, 96 $\|$ 1$\times$3, 96 $\|$ 3$\times$3, 64 \\
	\cline{4-4} \cline{7-9}	
	& \multicolumn{2}{c|}{} & 3$\times$1, 256 & & & 1$\times$1, 256 & 1$\times$1, 128 & 1$\times$1, 256\\
    \cline{2-9}
	& \multicolumn{8}{c|}{ReLU}\\
    \cline{2-9}
	& \multicolumn{8}{c|}{2$\times$2 maxpool, /2}\\
	\hline
	\textbf{conv4} & \multicolumn{2}{c|}{3$\times$3, 512} & 1$\times$3, 512 & 3$\times$1, 256 $\|$ 1$\times$3, 256 & 3$\times$1, 512 $\|$ 1$\times$3, 512 &\multicolumn{2}{c|}{3$\times$1, 256 $\|$ 1$\times$3, 256} & 3$\times$1, 192 $\|$ 1$\times$3, 192 $\|$ 3$\times$3, 128\\
	\cline{4-4} \cline{7-9}
	& \multicolumn{2}{c|}{} & 3$\times$1, 512 & & & 1$\times$1, 512 & 1$\times$1, 256 & 1$\times$1, 512\\
    \cline{2-9}
	& \multicolumn{8}{c|}{ReLU}\\
    \cline{2-9}
	& \multicolumn{2}{c|}{3$\times$3, 512} & 1$\times$3, 512 & 3$\times$1, 256 $\|$ 1$\times$3, 256 & 3$\times$1, 512 $\|$ 1$\times$3, 512 & \multicolumn{2}{c|}{3$\times$1, 256 $\|$ 1$\times$3, 256} & 3$\times$1, 192 $\|$ 1$\times$3, 192 $\|$ 3$\times$3, 128\\
	\cline{4-4} \cline{7-9}
	& \multicolumn{2}{c|}{} & 3$\times$1, 512 & & & 1$\times$1, 512 & 1$\times$1, 256 & 1$\times$1, 512\\
    \cline{2-9}
	& \multicolumn{8}{c|}{ReLU}\\
    \cline{2-9}
	& \multicolumn{8}{c|}{2$\times$2 maxpool, /2}\\
	\hline
    \textbf{conv5} & \multicolumn{2}{c|}{3$\times$3, 512} & 1$\times$3, 512 & 3$\times$1, 256 $\|$ 1$\times$3, 256 & 3$\times$1, 512 $\|$ 1$\times$3, 512 & \multicolumn{2}{c|}{3$\times$1, 256 $\|$ 1$\times$3, 256} & 3$\times$1, 192 $\|$ 1$\times$3, 192 $\|$ 3$\times$3, 128\\
    \cline{4-4} \cline{7-9}
    & \multicolumn{2}{c|}{} & 3$\times$1, 512 & & & 1$\times$1, 512 & 1$\times$1, 256 & 1$\times$1, 512\\
    \cline{2-9}
	& \multicolumn{8}{c|}{ReLU}\\
    \cline{2-9}
	& \multicolumn{2}{c|}{3$\times$3, 512} & 1$\times$3, 512 & 3$\times$1, 256 $\|$ 1$\times$3, 256 & 3$\times$1, 512 $\|$ 1$\times$3, 512 & \multicolumn{2}{c|}{3$\times$1, 256 $\|$ 1$\times$3, 256} & 3$\times$1, 192 $\|$ 1$\times$3, 192 $\|$ 3$\times$3, 128\\
	\cline{4-4} \cline{7-9}
    & \multicolumn{2}{c|}{} & 3$\times$1, 512 & & & 1$\times$1, 512 & 1$\times$1, 256 & 1$\times$1, 512\\
    \cline{2-9}
	& \multicolumn{8}{c|}{ReLU}\\
    \cline{2-9}
	& 2$\times$2 maxpool, /2 & \multicolumn{7}{c|}{global maxpool}\\
	\hline
	\textbf{fc6} & $7^2$ $\times$ 512 $\times$ 4096 & \multicolumn{7}{c|}{512 $\times$ 4096}\\
    \cline{2-9}
	& \multicolumn{8}{c|}{ReLU}\\
    \hline
	\textbf{fc7} & \multicolumn{8}{c|}{4096 $\times$ 4096}\\
    \cline{2-9}
	& \multicolumn{8}{c|}{ReLU}\\
    \hline
	\textbf{fc8} & \multicolumn{8}{c|}{4096 $\times$ 1000}\\
    \cline{2-9}
	& \multicolumn{8}{c|}{softmax}\\
    \hline
\end{tabular}
}
\caption[VGG Model Architectures.]{{\bf VGG Model Architectures}. Here ``3$\times$3, 32'' denotes 32 3$\times$3 filters, ``/2'' denotes stride 2, fc denotes fully-connected, and $\|$ denotes a concatenation within a composite layer.}
\label{table:vggarch}
\end{table*}
}

\end{landscape}